\documentclass[a4paper,10pt]{article}
\usepackage{amsmath}
\usepackage{amssymb}
\usepackage[dvipdfmx]{graphicx}
\usepackage{ifthen}
\usepackage{verbatim}
\usepackage{color}
\usepackage{slashbox}
\usepackage{mdwlist}
\usepackage{url}

\begin{document}

\bigskip
\bigskip
\begin{center}
{\Large \textbf{
	Hyperplane Arrangements and
 	Locality-Sensitive Hashing with Lift
}}
\end{center}

\vspace{10mm}
\begin{center}
Makiko Konoshima,
\footnote{E-mail: \texttt{makiko@jp.fujitsu.com}}
Yui Noma
\\
\bigskip
{\small 
\textit{
Software Systems Laboratories, FUJITSU LABORATORIES LTD.  1-1,\\
Kamikodanaka 4-chome, Nakahara-ku Kawasaki, 211-8588 Japan.\\}}
\end{center}

\vspace{10mm}
\begin{abstract}
Locality-sensitive hashing converts high-dimensional feature vectors,
 such as image and speech, into bit arrays and allows high-speed
 similarity calculation with the Hamming distance. 
There is a hashing scheme that
 maps feature vectors to bit arrays
 depending on the signs of the inner products between 
 feature vectors and the normal vectors of hyperplanes placed in the feature space.
This hashing can be seen as a discretization of the feature space by hyperplanes.
If labels for data are given,
 one can determine the hyperplanes by using learning algorithms.
However, many proposed learning methods do not consider the hyperplanes' offsets.
Not doing so decreases the number of partitioned regions,
 and the correlation between Hamming distances and Euclidean distances becomes small. 
In this paper, we propose a lift map
 that converts learning algorithms without the offsets
 to the ones that 
 take into account the offsets.
With this method,
 the learning methods without the offsets
 give the discretizations of spaces
 as if they take into account the offsets.
For the proposed method, we input several high-dimensional feature data sets
 and studied the relationship
 between the statistical characteristics of data,
 the number of hyperplanes, and the effect of the proposed method.

{\bf Keyword:}
 Locality-sensitive hashing, Arrangement, Higher-dimensional affine space,
 Similarity search, Precision, Recall, Error rate
 
\end{abstract}

\vspace{10mm}

\section{Introduction}

Unstructured data such as images and speech,
 which have complicated structures and are difficult to structure,
 account for 85\% or more of corporate data and are trending toward further increase.
Such data are used in various applications,
 such as the collection and comparison of similar data for various types of analysis,
 detection and reuse, and identity verification by using collected
 biological image data \cite{Seitai_JRP,Seitai_FujitsuPress}. 
The features of data used in such cases are high dimensional vectors,
 where the dimensions may reach as high as a few hundred to a few thousand,
 as higher search accuracy is required. 
For similarity search with high-dimensional feature vectors,
 many methods have been proposed as neighborhood search methods, such as
 KD-tree \cite{KDTree} and iDistance \cite{iDistance},
 which search for data that are a short distance from the query features. 
For a higher dimension,
 because of the curse of dimensionality, 
 it is shown that one cannot perform rapid similarity searches
 with the above methods~\cite{Weber1998DimensionCurse}.
Locality-sensitive hashing (LSH)
 \cite{LSH_IndykMotwani,LSH_RandomProjection,MLH,SIMBA_KN},
 which converts high-dimensional feature vectors into
 bit arrays and allows high-speed similarity calculation with
 the Hamming distance method, is attracting attention as a technology
 capable of solving such problems and can also benefit high-speed,
 large-scale data searching.

One of the most investigated schemes for locality-sensitive hashing
 is the random projection~\cite{LSH_RandomProjection}.
In that method,
 one considers a feature space as a vector space
 and partitions the space by using hyperplanes crossing the origin.
The data are assigned 0 or 1 in accordance with the signs of the inner
 products with respect to the normal vectors of the hyperplanes.
When the number of bits becomes large,
 the correlation between Hamming distances and the angles of data pairs
 becomes strong.
However, in some cases,
 a feature space is an affine space.
In such cases,
 because there are no special points in the feature space,
 the angles are not appropriate characteristics for similarities.
One of the appropriate characteristics for similarities
 are distances.
Hence, we need a hashing scheme in which
 Hamming distances approximate Euclidean (L2) distances up to a factor.
Obviously, this is done by
 discretizing the space with hyperplanes that have arbitrary directions and positions.

\begin{figure}[tbp]
	\begin{center}
	\includegraphics[scale=0.05]{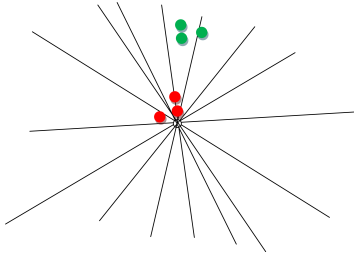}
	\end{center}
	\caption{Effects of no offset on space partitioning.}
	\label{fig_EffectNoOffset}
\end{figure}

Although a data space is a high-dimensional affine space,
 in the following, we see the data space 
 as a vector space with one point
 in the space fixed to serve as the origin,
 which is, for example, the mean of sample data.
A hyperplane is described in terms of a normal vector and an offset,
 which is the minimum distance between the hyperplane and the origin.
To consider partitions of space by hyperplanes
 as a discretization of space,
 one has to take account of the offsets of hyperplanes
 because,
 if the offsets are not considered, there will be fewer regions partitioned
 by hyperplanes than in a case that includes offsets. 
In addition, increasing the number of hyperplanes for partitioning to improve
 the accuracy of the distance calculation results in partitioning into
 more minute spaces near the origin, which is the point of the intersection of all hyperplanes. 
This leads to a greater difference in the partitioning sizes
 from spaces that are distant from the origin.
Fig.~\ref{fig_EffectNoOffset} shows this effect.
Therefore, we consider partitions of space by hyperplanes with offsets.

In the case where labels are attached to data,
 L2 distances may not be appropriate dissimilarities.
In that case,
 one may determines the hyperplanes with machine learning
 so that the Hamming distances between same-labeled data become small. 
When one determine the hyperplane normal vectors by direct learning,
 the algorithm will be complicated if the offsets are also taken into account.

We propose a method for solving the problem described above. 
The proposed methodology does not take offsets into consideration
 but rather uses lifting to prevent the concentration of intersections
 on the origin and increases the number of space partitions
 to produce an effect equivalent to offsetting.
The benefits of the proposed method include easy application to
 algorithms that do not take offsets into consideration and,
 for future study of learning algorithms,
 the elimination of the need to consider offsets. 
This may possibly simplify the study of algorithms.

We applied the proposed method to the random projection method
 of hashing,
 minimal loss hashing~\cite{MLH},
 which does not take into consideration offsets as the algorithm,
 and LSH with margin based feature selection(S-LSH)~\cite{SIMBA_KN}. 
With reference to these, we processed the following 
 data, which have different statistical characteristics:
 a LabelMe natural image database~\cite{LabelMe}, MNIST handwritten digit database~\cite{MNIST},
  mel frequency cepstral coefficient (MFCC) features of speech~\cite{SpeechMFCC},
 and features
 resulting from Fourier transform of fingerprint images~\cite{Finger_power}. 
The effect of the proposed method and the relationship with the number of bits are studied.
The results are reported in this paper.

\section{Background and related work}

\subsection{Locality-sensitive hashing with hyperplanes}

Let $V$ be a data space. 
A hyperplane $H_{(\vec{n}, b)}$ can be described in terms of
 a normal vector $\vec{n}$, and an offset from the origin, $-b$:
\begin{eqnarray}
	H_{(\vec{n}, b)} = \left\{
		\vec{x} \in V | \vec{n} \cdot \vec{x} + b = 0
	\right\} .
\end{eqnarray}

Many proposals have been made regarding the method for determining hyperplanes.
One typical method is the random projection (LSH)~\cite{LSH_RandomProjection},
 in which hyperplanes are generated by random numbers. 
This method approximates the accuracy of the original cosine similarities
 when the number of hyperplanes is sufficiently large. 
In addition to the random projection, various methods for determining hyperplanes
 by learning have been proposed. 
Two of the methods that we consider in this paper are
 minimal loss hashing (MLH)~\cite{MLH} 
 and LSH with margin based feature selection (S-LSH)~\cite{SIMBA_KN}).
These methods are optimization based learning.


With methods such as MLH, all hyperplanes cross the origin, and offsets are not
 taken into consideration. 
Therefore, we cannot apply these methods naively to learn the discretization of a space.

\subsection{Importance of hyperplane offsets}

\begin{figure}[tbp]
	\begin{center}
	\includegraphics[scale=0.25]{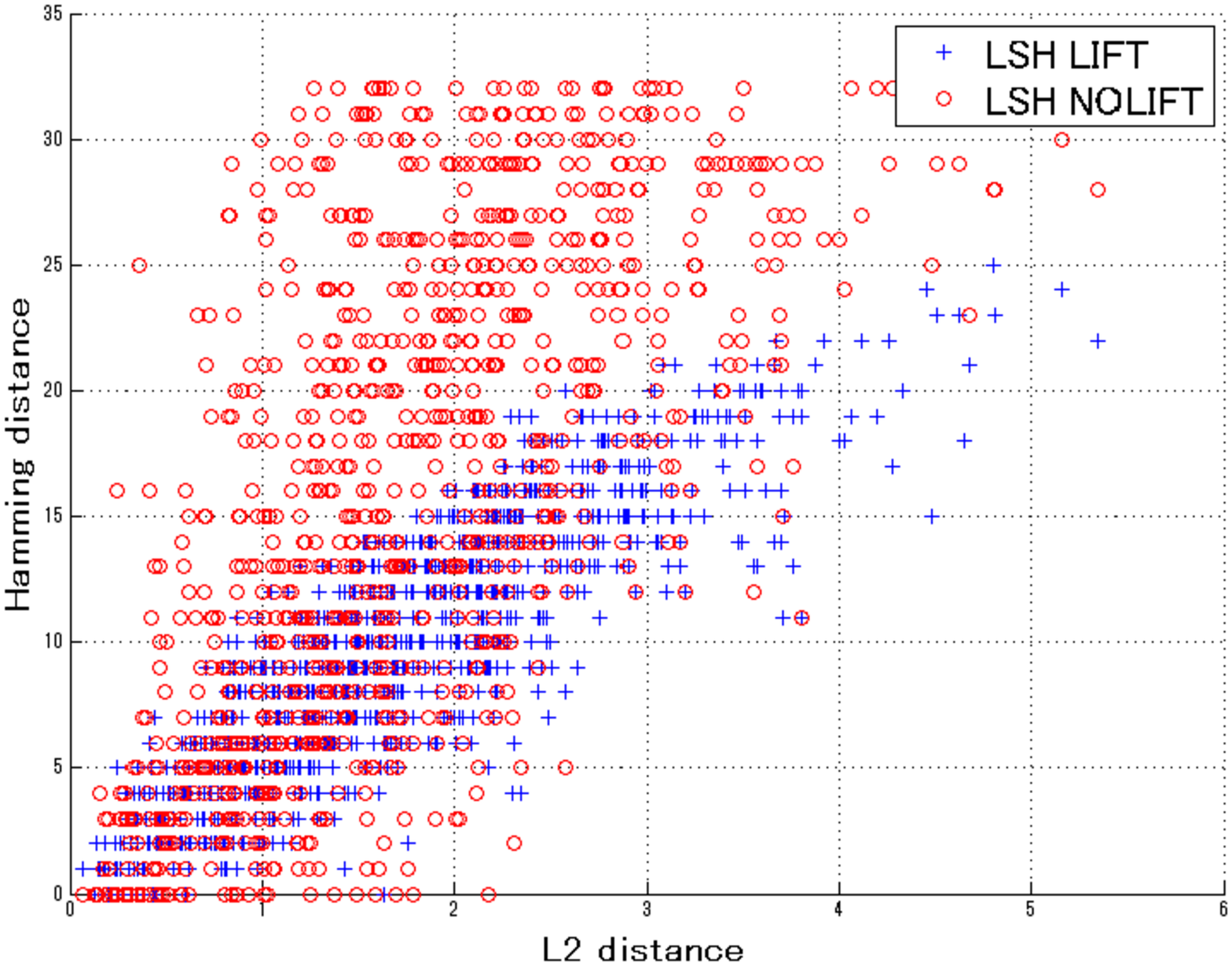}
	\includegraphics[scale=0.25]{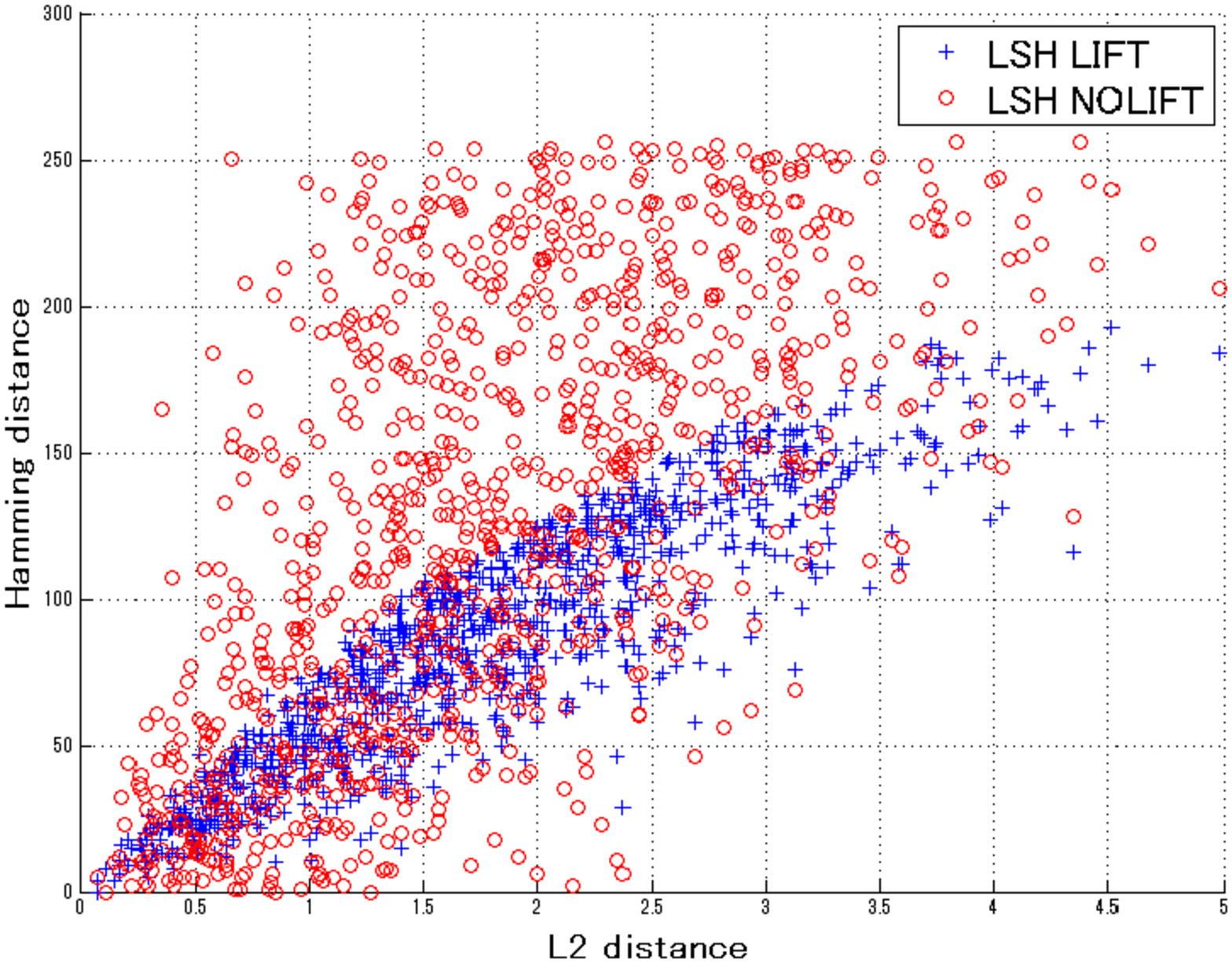}
	\end{center}
	\caption{Correlation between L2 distances and Hamming distances hashed by hyperplanes
		with offset (represented by +)
		and without offset (represented by circles).
		The number of bits is 32 (left) and 256 (right).}
	\label{fig_CorrelationL2Hamming}
\end{figure}

In this subsection, we describe the importance of
 hyperplane offsets from the perspective of the accuracy
 of the approximation of an L2 distance relative to the Hamming distance. 
Fig.~\ref{fig_CorrelationL2Hamming} shows the relations
 between L2 distances and Hamming distances with LSH. 
The figures were made by
 picking data pairs on a two dimensional space randomly,
 and calculating their L2 distances and their Hamming distances
 hashed by hyperplanes with offset
 and
 those crossing the origin.
From Fig.~\ref{fig_CorrelationL2Hamming}, one can see that
 Hamming distances hashed
 by hyperplanes with offset
 approximate L2 distances with higher accuracy regardless of the number of bits.

\section{Hashing with lift}

The proposed method maps all data into a space dimensionally
 increased by one and, in that space,
 performs learning hyperplanes crossing the origin. 
An arbitrary learning algorithm for hyperplanes can be used to accomplish this.

This proposed method is a higher dimensional analog of
 the horizontal lift of curves on the base space
 with a flat connection on a trivial fiber bundle~\cite{Nakahara_GeometryTopologyPhysics}.

We explain the lift proposed in this paper. 
Let $N$ be the dimension of data space $V$. 
Consider a $N+1$ dimensional space $W$ and the following embedding.
\begin{eqnarray}
	f : V &\rightarrow& W , \\
	f(x_1, x_2, \cdots, x_N) &=&
		(x_1, x_2, \cdots, x_N, 1).
\end{eqnarray}
The $(N+1)$th component of $W$ is referred to as $z$.
If we consider a naive extension of $V$ by adding an extra dimension,
 it is natural to identify $V$ as a $z=0$ hyperplane.
Hence, the embedding $f$ is a lift of $V$
 from the $z=0$ hyperplane to the $z=1$ hyperplane.
This is the reason we call $f$ "lift."
In the following, we identify $V$ and the $z=1$ hyperplane.

Let us consider a hyperplane that crosses the origin of $W$. 
This hyperplane crosses the $z=1$ plane when the normal vector
 is not parallel to the $z$ axis. 
The dimension of the common set is $N-1$. 
The common set does not necessarily cross the origin of $V$. 
That is, a hyperplane that crosses the origin of $W$ is mapped to
 a hyperplane with an offset in $V$. 
The degree of freedom of a hyperplane crossing the origin of
 $W$ is $N$, and the degree of freedom of a hyperplane with an offset
 in $V$ is also $(N-1)+1$.
Note that the degree of freedom of a normal vector of a hyperplane in an $n$ dimensional space is $n-1$.
To consider a specific map between hyperplanes,
 let $\tilde{\vec{n}} = (\vec{n}, w)$ be a normal vector of a hyperplane that crosses
 the origin of $W$. 
If a point on the $z=1$ plane in $W$ is denoted as $\tilde{\vec{x}} = (\vec{x}, 1)$, 
the common set of the hyperplane and the $z=1$ plane is:
\begin{eqnarray}
	0 = \tilde{\vec{n}} \cdot \tilde{\vec{x}}
	= (\vec{n}, w) \cdot (\vec{x}, 1) = \vec{n}\cdot \vec{x} + w .
\end{eqnarray}
This is the equation for a hyperplane with an offset in $V$.

\section{Experiments}

We verified the effect of the proposed method based on LSH,
 MLH, and S-LSH. 
For implementation, we took the following steps. 
The 1st to $(N + 1)$th components of a normal vector of a hyperplane that
 crosses the origin of $W$ were sampled from the standard normal distribution.
They were normalized so that a unit vector was composed of only the 1st to $N$th components. 
If seen as a hyperplane in $V$, this initialization is equivalent to
 the one in which
 normal vectors are sampled from the $N$-dimensional standard normal distribution
 and offsets are sampled from the standard normal distribution.

The following explains why we selected the normal distribution for offsets
 rather than a uniform distribution. 
For learning, normalization was applied before a principal component
 analysis of data so that the mean of the respective components of a feature
 was 0 and the standard deviation was 1.
In spite of the normalization,
 existing data and unknown data are not necessarily included
 in a compact set in general. 
For that reason, using hyperplanes with arbitrary directions and offsets
 for separating data with different labels is expected to offer higher separation
 performance for certain types of data. 
Let $p$ stand for the point that provides the shortest distance from the origin on a hyperplane. 
When hyperplane offsets are sampled from a uniform distribution
 with the minimum value 0 and the maximum value $r$,
 point p can exist only within the sphere of radius r. 
Accordingly, the separation performance is low for data with different labels
 outside the sphere with radius $r$. 
Meanwhile, sampling hyperplane offsets from the standard normal distribution allows
 point $p$ to exist in any position in the space,
 and the probability is in line with the normalized data distribution.

To evaluate the performance of the proposed method, 
 we performed calculations by using the following data types:
 natural image data (hereafter "LabelMe"), handwritten digit data (hereafter "MNIST"),
 speech feature data, and fingerprint image data. 
The learning algorithms were LSH, MLH, and S-LSH with hyperplanes that cross the origin,
 and LSH, MLH, and S-LSH with lifting applied. 
We studied precision and recall, which are indicators of search accuracy
 and are defined as follows.
\begin{eqnarray}
	\mbox{Precision} &:=& \frac{
				\begin{array}{l}
					\mbox{\small Number of data of the same label} \\
					\mbox{\small \: as the query acquired by search}
				\end{array}}
			{\mbox{\small Number of data acquired by search}} ,\\
	\mbox{Recall} &:=& \frac{
				\begin{array}{l}
					\mbox{\small Number of data of the same label}\\
					\mbox{\small \: as the query acquired by search}
				\end{array}}
			{\begin{array}{l}
				\mbox{\small Number of data of the same label}\\
				\mbox{\small \: as the query of data searched}
			\end{array}} .
\end{eqnarray}

The above data types have different total numbers of labels and different numbers of data
 included in the same label. 
Search results are composed of high-ranked data sorted in ascending order of
 the Euclidean distances or the Hamming distances with respect to a query.
Accordingly, the rate of search results acquired from the data searched is
 defined as the acquisition:
\begin{eqnarray}
	\mbox{Acquisition} &:=& \frac{\mbox{Number of data acquired by search}}
					{\mbox{Total number of data searched}}  .
\end{eqnarray}

We will explain the experimental procedure. 
Assuming similarity search,
 we classified data into the following three kinds:
 data for learning,
 data in a database,
 and data used for query.
These three sets are mutually disjoint.
The data are normalized so that the mean of the respective components
 is 0 and the standard deviation is 1.
Principal component analysis (PCA) was performed on the normalized learning data. 
The data were projected to the subspace where the cumulative contribution ratio
 exceeded or equaled 80\%. 
By using the projected data for learning, the hyperplanes were learned.
Since labeling between data depends on the data types,
 they are described in the explanation of each experiment. 
The numbers of bits were 32, 64, 128, 256, 512, and 1,024.
For S-LSH, the initial number of bits was $10^4$, and the number of iterations for learning was $10^4$.
For MLH, the number of iterations for learning was $10^4$.

\subsection{Experiments on LabelMe}

We used 512-dimensional Gist features~\cite{GIST} of
 the natural image data of the LabelMe data set~\cite{LabelMe},
 which is used in the literature~\cite{512dimGISTdata}. 
The labeling was done as follows. We normalized the data,
 calculated a non-similarity matrix by using L2 distance,
 sorted the matrix in ascending order,
 and determined that the top 1\% pairs have the same labels.
The total number of data was 22,000. 
The number of data was 11,000 for learning, 5,500 for query, 5,500 for data searched. 
The dimension of the subspace to which data were projected
 was 28.

\begin{figure}[tb]
 	\begin{center}
  	\includegraphics[scale=0.40]{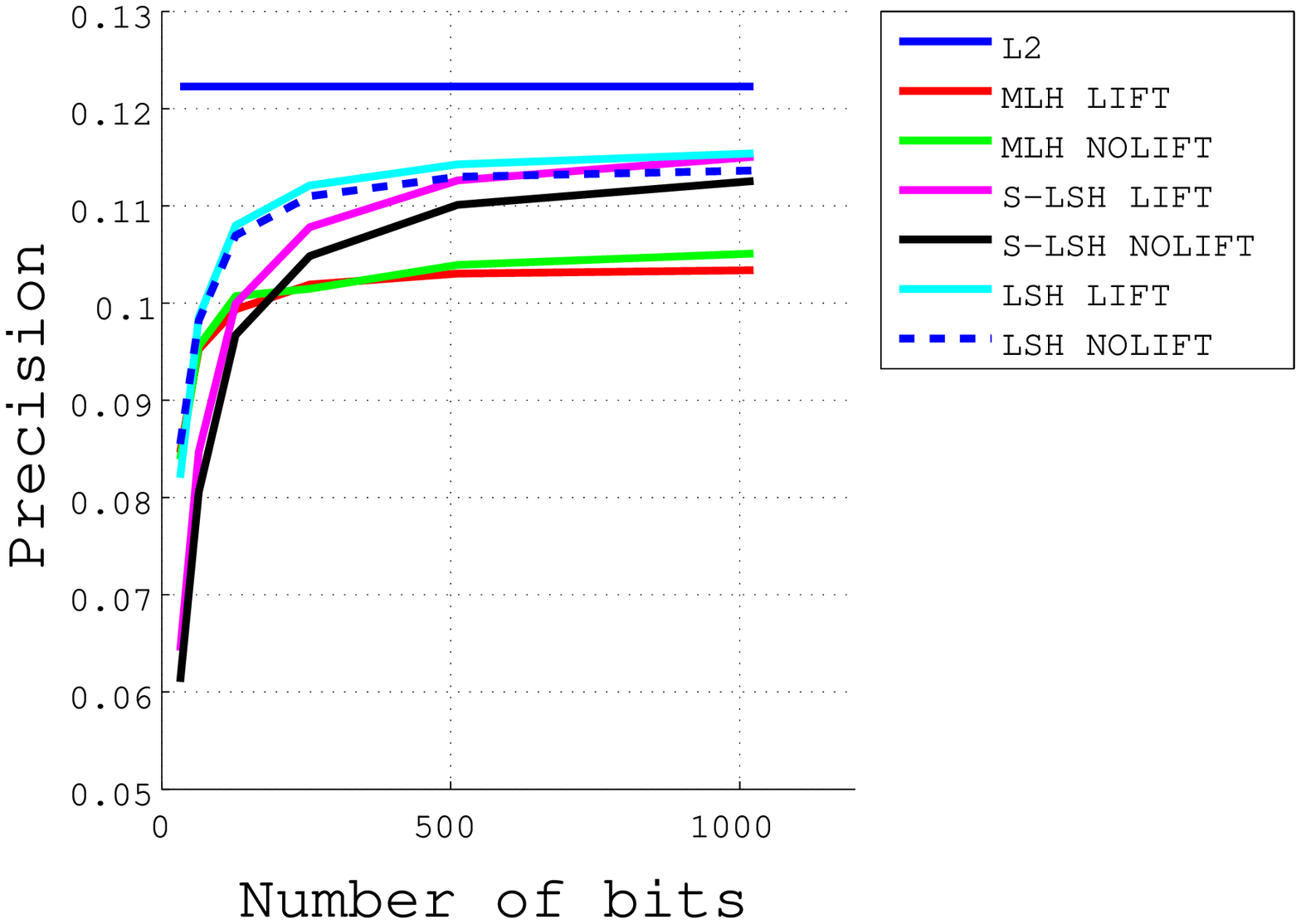}
	\includegraphics[scale=0.40]{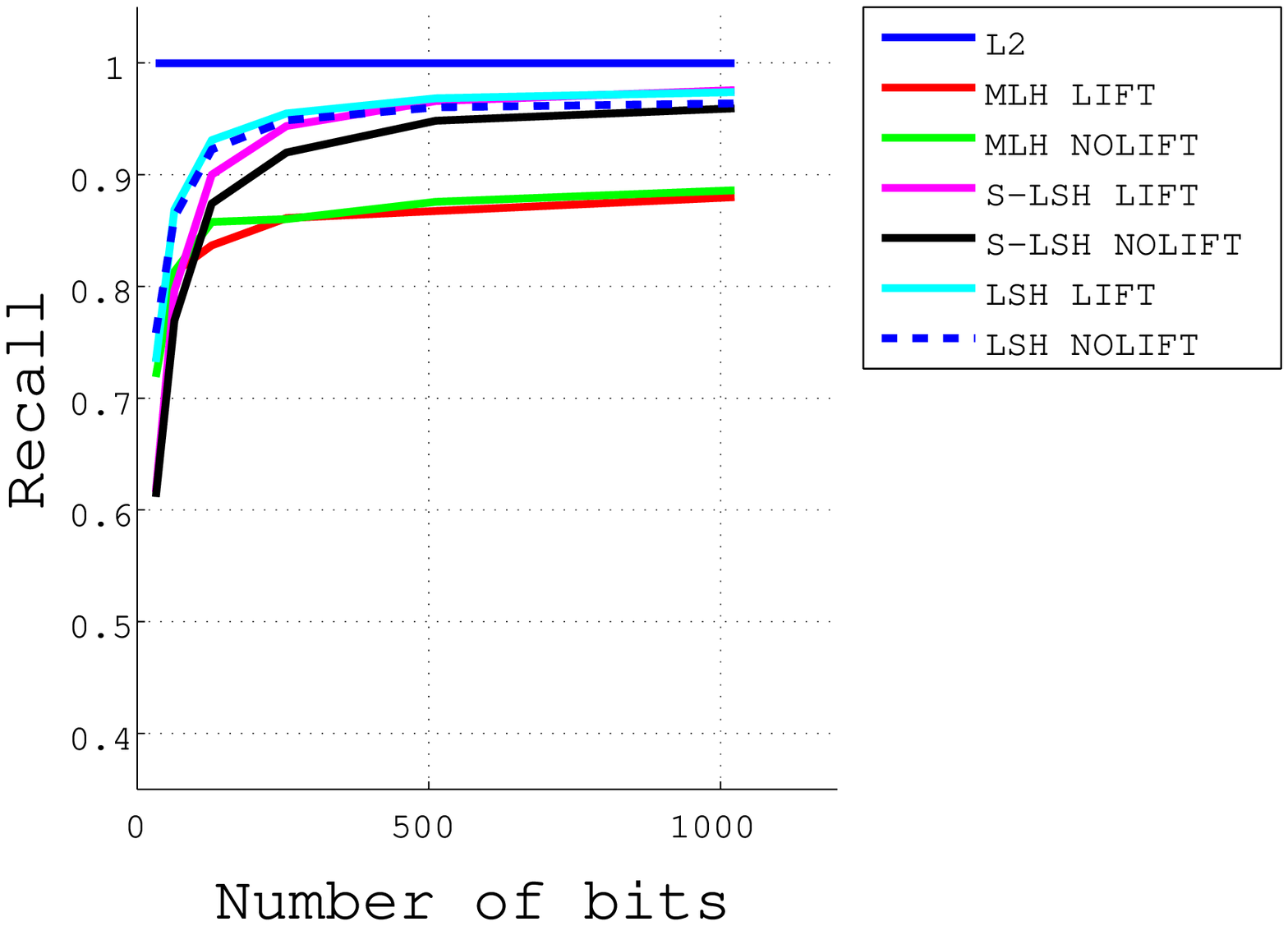}
 	\end{center}
 	\caption{
		LabelMe: Curves of precision (left) and recall (right) versus the number of bits for L2, MLH, S-LSH, and LSH with or without lift.
		}
 	\label{fig_LabelMePrecision}
\end{figure}

Fig.~\ref{fig_LabelMePrecision} 
 show graphs of the dependency
 of the precision and recall on the number of bits with the acquisition fixed at $0.1$. 
We can observe performance improvements of the proposed method for LSH and S-LSH.
Since the learning performance of MLH was not good,
 we could not observe the effect of lifting for MLH.

\subsection{Experiments on MNIST}

MNIST is a set of 8-bit grayscale images of handwritten digits, 0 to 9,
 consisting of $28 \times 28$ pixels per digit,
 which are individually assigned labels of 0 to 9 \cite{MNIST}. 
The number of data is 60,000 for learning data,
 5,000 for query data, 
 and 5,000 for data searched.

We used this image data as features to study precision and recall.
The dimension after the reduction was 149.

\begin{figure}[tb]
	\begin{center}
	\includegraphics[scale=0.40]{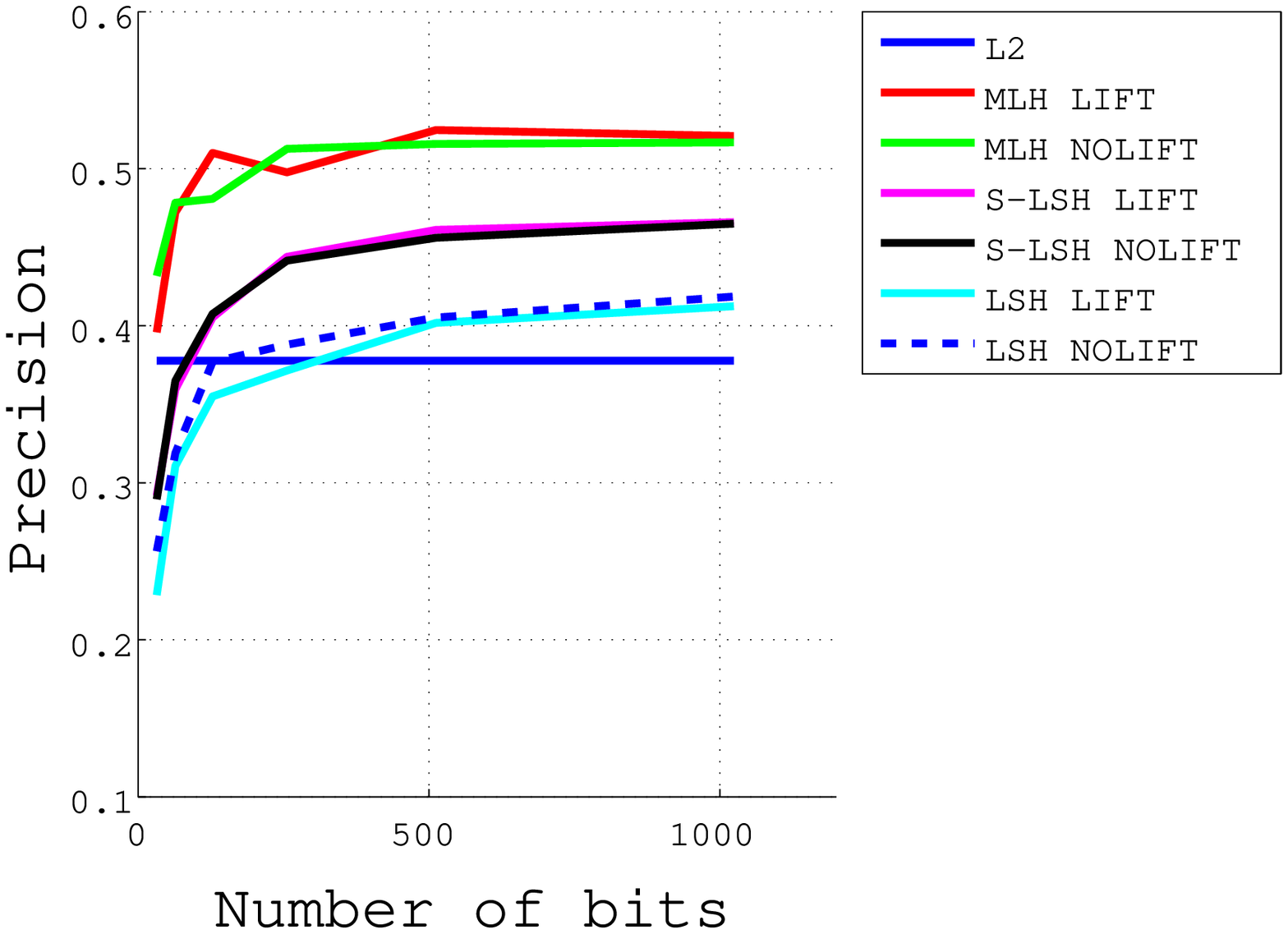}
	\includegraphics[scale=0.40]{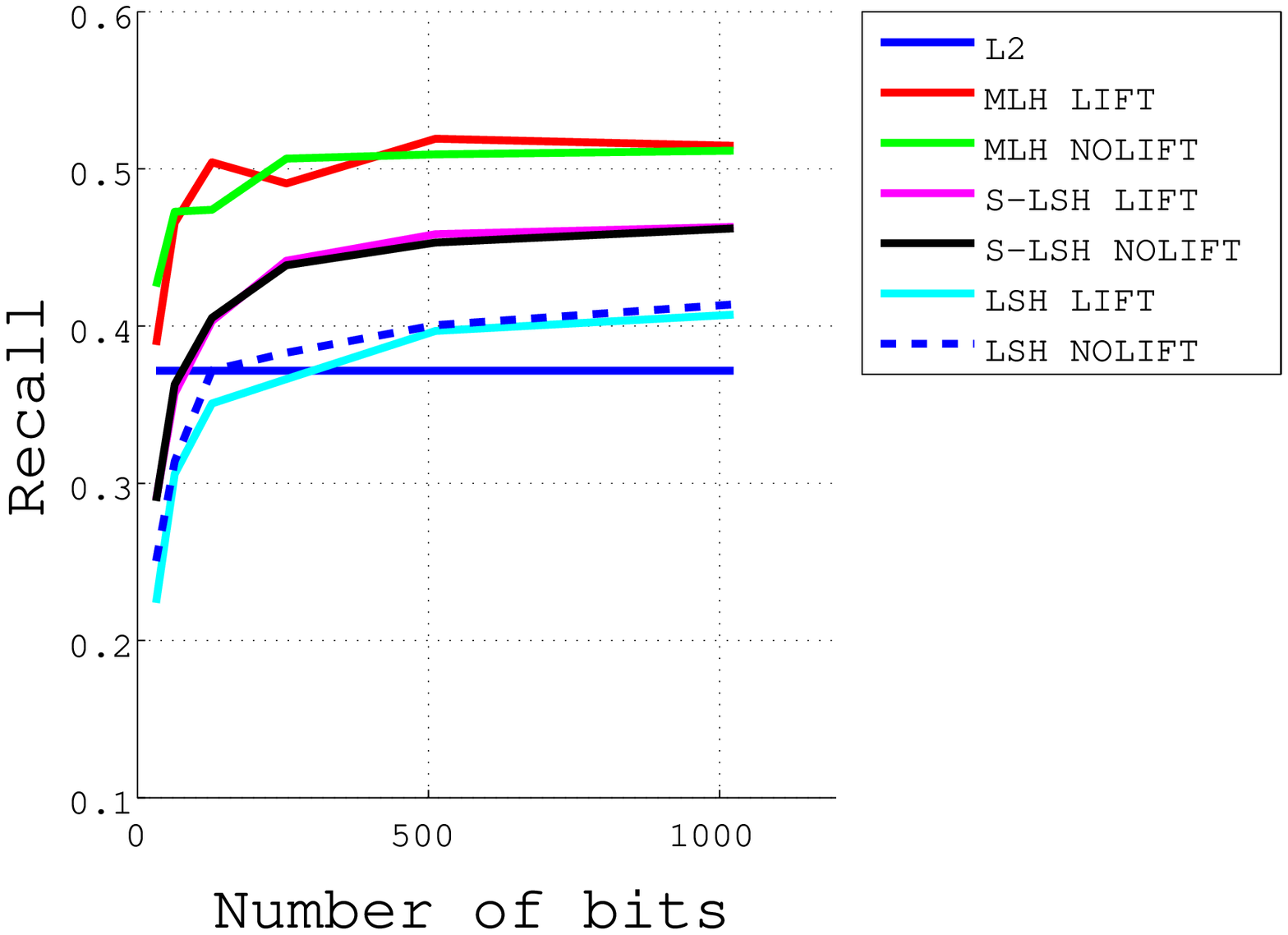}
	\end{center}
	\caption{
		MNIST: Curves of precision (left) and recall (right) versus the number of bits for L2, MLH, S-LSH, and LSH with or without lift.
		}
	\label{fig_MNISTPrecision}
\end{figure}

MNIST offers ten types of handwritten digit data.
For this reason,
 graphs of the dependency of the precision and recall on the number of bits
 with the acquisition fixed at 0.1 are shown in
 Fig.~\ref{fig_MNISTPrecision}.
We can see that the proposed method reduced the performance for LSH. 
The bad effects of lifting were reduced for MLH and S-LSH due to their effects of learning. 

\subsection{Experiments on speech features}

For the experiment, we used an internet relay broadcast of
 a three-hour-long local government council meeting~\cite{KawasakiCouncil}.
Because speech data are temporally continuous, we used 200-dimensional
 MFCC data obtained by using overlapping window functions as features. 
For the queries, we used speech sounds separately obtained.

For supervised learning of speech, speech content (text) is usually used as labels.
For this study, however, we assumed a collection of data with similar features,
 and those features with the top 0.1\% shortest Euclidean distances
 from the queries were regarded to be in the same class for implementing learning and evaluation.

For learning, 192,875 pieces of MFCC data obtained from 378 pieces of speech data were used.
The number of data was 1,815 for query data and 192,683 for data searched. 
The number of dimensions after the reduction
 was 30.

\begin{figure}[tb]
	\begin{center}
	\includegraphics[scale=0.40]{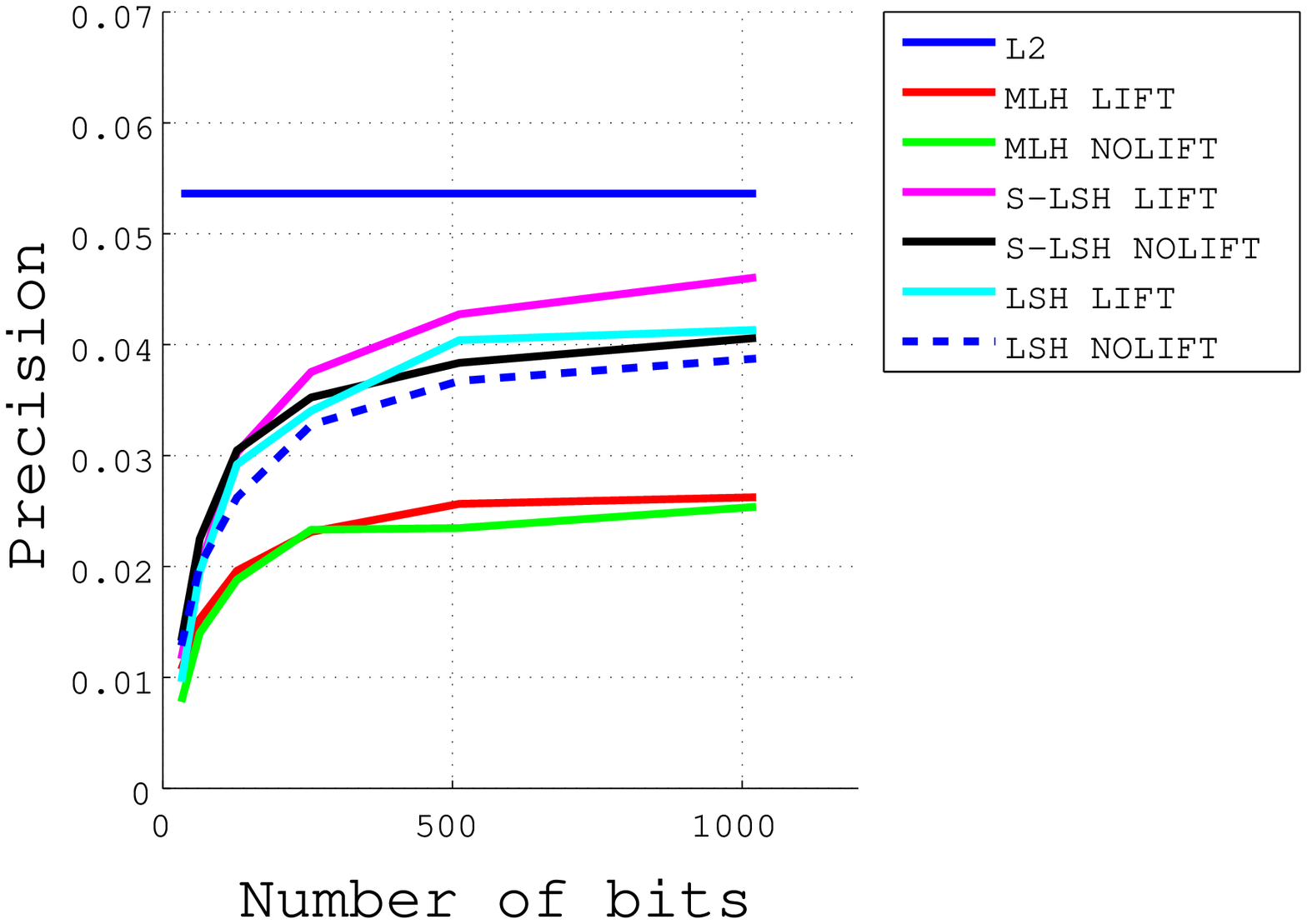}
	\includegraphics[scale=0.40]{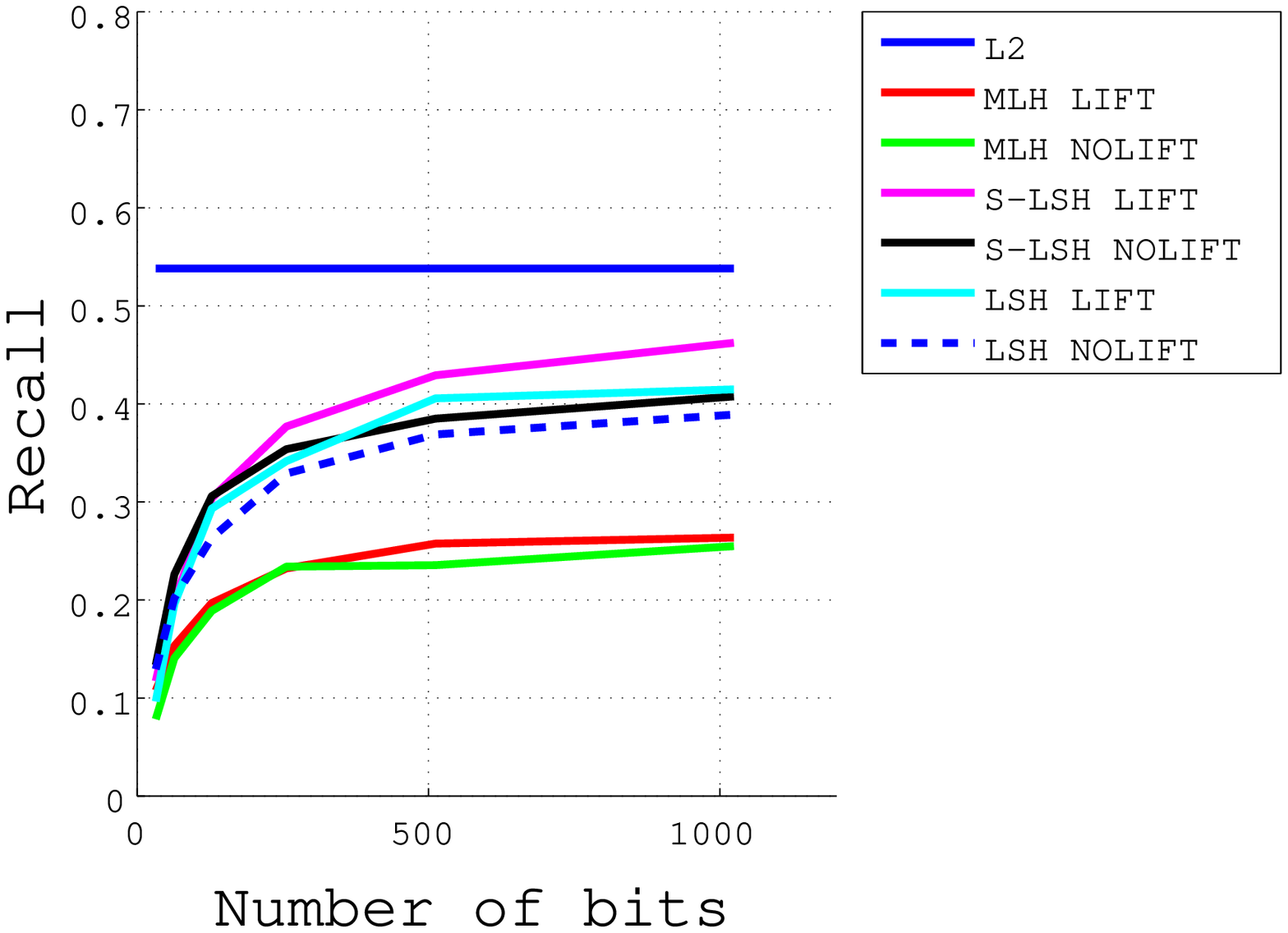}
	\end{center}
	\caption{
		Speech features: Curves of precision (left) and recall (right) versus the number of bits for L2, MLH, S-LSH, and LSH with or without lift.
		}
	\label{fig_SpeechPrecision}
\end{figure}

Fig.~\ref{fig_SpeechPrecision}
 show graphs of the dependency of the precision and recall on the number of bits
 with the acquisition fixed at 0.01. 
They indicate that lifting had a tendency to be more effective with a larger number of bits
 with both LSH and S-LSH. 
Since the learning performance of MLH was not good,
 we could not observe the effect of lifting for MLH.

\subsection{Experiments on fingerprint images}

For the experiment, we collected fingerprint images with a fingerprint reader. 
The feature vectors of the fingerprint images were 4096-dimensional floating point data,
 which were the power spectrum of the images.

The following describes how fingerprint image data were collected. 
Twelve sets of image data were collected for each of the right and left
 index, middle, and ring fingers
 of 1032 people.
Those obtained in poor collection conditions
 were not used, and the same label was shared by up to 12 pieces of data.
Labels were determined for each finger of an individual at the time of collection,
 which means that labels could be automatically assigned. 
Because the biological features of the respective fingers,
 as well as the right and left hands, are mutually independent even for one person,
 the number of labels was 6,192.

We used the data of approximately 25\% of all persons
 for learning.
The number of data was 9,906 for learning,
 19,932 for query data, and 12,138 for data searched. 
The number of dimensions after the reduction
 was 276.

Compared with LabelMe, MNIST, and speech feature, the data of the fingerprint features
 had some unique characteristics.
The number of same-labeled data was quite small.
The data with the same label belonged to the same individual.
When fingerprint images are used to identify people,
 it is important whether or not a specific individual is included
 in a search result.
On the basis of this,
 we defined the error rate as the probability that the data obtained
 by search do not include the same label as the queries:
\begin{eqnarray}
	\mbox{Error\: rate} := 
			\frac{
				\begin{array}{l}
					\mbox{\small Number of query data whose labels are not}\\
					\mbox{\small \: assigned to the search result data}
				\end{array}}
			{\mbox{Total number of query data}}.
\end{eqnarray}

\begin{figure}[tb]
	\begin{center}
	\includegraphics[scale=0.40]{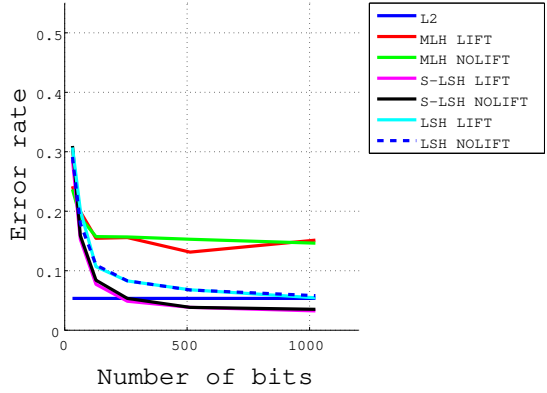}
	\end{center}
	\caption{
		Fingerprint features:
		Error rate versus number of bits for L2, MLH, S-LSH, and LSH with or without lift.
		}
	\label{fig_FingerError}
\end{figure}

Graphs of the dependency of the error rate on the number of bits with the acquisition
 specified as 0.1 are shown in Fig.~\ref{fig_FingerError}.

Fig.~\ref{fig_FingerError} indicates that
 almost no effect of lifting was observed on the fingerprint images. 

\section{Discussion}

We list the result in Table \ref{table_EffectOfLift}.
\begin{table}[tb]
	\caption{Effect of lift. Symbols stand for
		improved $(\circ)$, cannot be observed $(\bullet)$, and deterioration $(\times)$.}
	\begin{center}
	\begin{tabular}{c|r|r|r|r}
	\hline
	\hline
	\backslashbox{Method}{Data set} & LabelMe & MNIST & Speech & Fingerprint \\
	\hline
	MLH    & $\bullet$ & $\bullet$ & $\bullet$ & $\bullet$ \\
	S-LSH  & $\circ$   & $\bullet$  & $\circ$   & $\bullet$ \\
	LSH    & $\circ$   & $\times$  & $\circ$   & $\bullet$ \\
	\hline
	\end{tabular}
	\label{table_EffectOfLift}
	\end{center}
\end{table}
From the results of these experiments,
 we conclude that the performance improvement effect of the proposed method exists. 
It is believed that this effect depends highly on the statistical
 features of data sets and the features of labels.

We believe that the following statistical features of data sets
 maximize the performance improvement effect of lifting.
\begin{enumerate}
	\item The number of bits is sufficiently larger than the dimension of the space
		to which data are projected.
		\label{enum_Reason1}
	\item The standard deviations of the principal component directions are small.
		\label{enum_Reason2}
\suspend{enumerate}
The following features of labels are considered
 to maximize the performance improvement effect of lifting. 
Let us denote data sets whose elements have the same label by the labeled sets.
\resume{enumerate}
	\item Many labeled sets surround the origin.
		Therefore, one could not divide the labeled sets by hyperplanes crossing the origin.
		\label{enum_Reason3}
	\item Many labeled sets are separated along the radial direction and not separated
		along the angular direction.
		\label{enum_Reason4}
\end{enumerate}
In the following, we explain the effects of these features
 to the performance-improvement effect of lifting.

The effect of No.~\ref{enum_Reason1} is the following. 
When the number of bits is smaller than or equal to the dimension of the feature space,
 the number of regions separated by the hyperplanes crossing or being apart from the origin
 is $2^{\mbox{number of bits}}$,
 which is the maximum number of regions labeled by the bit array.
However when the number of bits is much higher than the dimension of the feature space,
 the number of regions separated by the hyperplanes crossing the origin
 is $O((\mbox{number of bits})^{\mbox{dimension}-1}  )$. 
It is much smaller than the maximum number.
This implies that
 the discretization performance of hashing is bad at a lower dimension and
 higher number of bits. 
Since hashing with lift is equivalent to the hashing with offset hyperplanes,
 the number of separated regions is $O((\mbox{number of bits})^{\mbox{dimension}}  )$
 for hashing with lift. 
Therefore, lifting improves
 the discretization performance of hashing  at a lower dimension and
 higher number of bits. 
The experimental results, which express this effect, are LSH for speech features
 (see Fig.~\ref{fig_LabelMePrecision} for reference).

\begin{figure}[tbp]
 	\begin{center}
 	\includegraphics[scale=0.40]{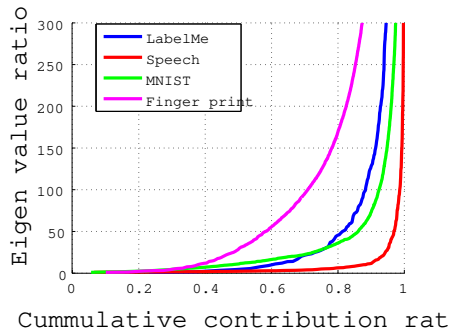}
 	\end{center}
 	\caption{Ratio between 1st eigenvalue and n-th eigenvalue}
 	\label{fig_STDRatio}
 \end{figure}

The effect of No.~\ref{enum_Reason2} is the following. 
Each piece of feature data are normalized before principal component analysis
 of the data so that the mean of the respective components of the feature
 is 0 and the standard deviation is 1. 
However, the standard deviations of the principal component directions are 
 significantly different from one due to the curse of dimensionality. 
Therefore, it does not fit to the distribution of data that the offsets are
 sampled from the standard normal distribution. 
Fig.~\ref{fig_STDRatio} shows the ratio between the 1st eigenvalue 
 and n-th eigenvalue of the PCA for various data types.
We list the ratio between the 1st and n-th eigenvalues,
 where n is determined such that the cumulative contribution ratio exceeds 80\%,
 which for the LabelMe, MNIST, speech features, and fingerprint features
 is respectively $46.1, 36.2, 6.5$, and $169.8$.
This means that the distribution of data is not spherical but elliptical.
In particular, while the deviation of the speech features,
 for which lifting works well, was $O(10^0)$,
 the deviation of the fingerprint features,
 for which lifting does not work well, was $O(10^2)$.

The effect of No.~\ref{enum_Reason3} appears notably for the case of the speech features. 
The opposite nature appears for the case of MNIST.
Because the speech features are labeled by using L2 distances,
 there are many labeled sets that surround the origin. 
It is hard to show directly
 where labeled sets surround the origin
 in the speech features and MNIST.
Hence,
 we calculated the means of the absolute value
 of the resulting offset of the hyperplanes
 for S-LSH, whose learning effect exists for both data types,
 and LSH, which is the base of S-LSH.
If many labeled sets surround the origin,
 the mean for S-LSH must be smaller than that for LSH.
In the case of speech features,
 the values were 0.0128 and 0.0134 for LSH and S-LSH, respectively. 
Conversely, in the case of the MNIST data set, the means were 0.0123 and 0.0048
 for LSH and S-LSH, respectively. 
This implies that for the case of MNIST
 many labeled sets did not surround the origin,
 and
 it was adequate to separate the labeled set
 by using hyperplanes crossing the origin.

The effect of No.~\ref{enum_Reason4} is obvious. 
However, we could not prepare a data set which possesses such property.

\section{Conclusion}

In this paper, we proposed a lifting method that increases the number
 of regions separated by hyperplanes and improves performance
 for locality-sensitive hashing with hyperplanes crossing the origin. 
By using this method,
 one can obtain learning methods for distances,
 which take into account the offsets of hyperplanes,
 from any learning method for angles,
 which does not take into account the offsets of hyperplanes.
We applied the proposed method to various data sets that possess
 different statistical properties. 
As a result, we found that the proposed method works well
 for data sets with the following characteristics:
 the labeled sets surround the origin,
 the data are distributed spherically,
 and the dimension of the features
 is much smaller than the number of bits.


\bibliographystyle{unsrt}
\bibliography{BibForHashing}

\end{document}